\documentclass[10pt,twocolumn,letterpaper]{article}

\usepackage{cvpr}
\usepackage{times}
\usepackage{epsfig}
\usepackage{graphicx}
\usepackage{amsmath}
\usepackage{amssymb}
\usepackage{booktabs}

\usepackage[pagebackref=true,breaklinks=true,letterpaper=true,colorlinks,bookmarks=false]{hyperref}

\cvprfinalcopy 

\def\cvprPaperID{2384} 

\ifcvprfinal\fi
\begin{document}

\title{Generalized Deep Image to Image Regression}

\author{Venkataraman Santhanam, Vlad I. Morariu, Larry S. Davis\\
UMIACS\\
University of Maryland, College Park\\
{\tt\small [venkai,morariu,lsd]@umiacs.umd.edu}
%
}

\maketitle

\begin{abstract}
   We present a Deep Convolutional Neural Network architecture which serves as a generic image-to-image regressor that can be trained end-to-end without any further machinery. Our proposed architecture: the Recursively Branched Deconvolutional Network (RBDN) develops a cheap multi-context image representation very early on using an efficient recursive branching scheme with extensive parameter sharing and learnable upsampling. This multi-context representation is subjected to a highly non-linear locality preserving transformation by the remainder of our network comprising of a series of convolutions/deconvolutions without any spatial downsampling. The RBDN architecture is fully convolutional and can handle variable sized images during inference. We provide qualitative/quantitative results on $3$ diverse tasks: relighting, denoising and colorization and show that our proposed RBDN architecture obtains comparable results to the state-of-the-art on each of these tasks when used off-the-shelf without any post processing or task-specific architectural modifications.  
\end{abstract}

\section{Introduction}

Over the last few years, generic deep convolutional neural network (DCNN) architectures such as variants of VGG~\cite{refvgg} and ResNet~\cite{refresnet} have been immensely successful in tackling a diverse range of classification problems and achieve state-of-the-art performance on most benchmarks when used out of the box. The key feature of these architectures is an extremely high model capacity along with a robustness to minor unwanted (\eg translational/rotational/illumination) variations. Given suitable training data, such models can be discriminatively trained in a reliable end-to-end fashion. However, since classification tasks only require a single (potentially multi-variate) class label corresponding to the entire image, early architectures focused solely on developing strong global image features.

\begin{figure}[htpb]
\begin{center}
\includegraphics[width=\linewidth]{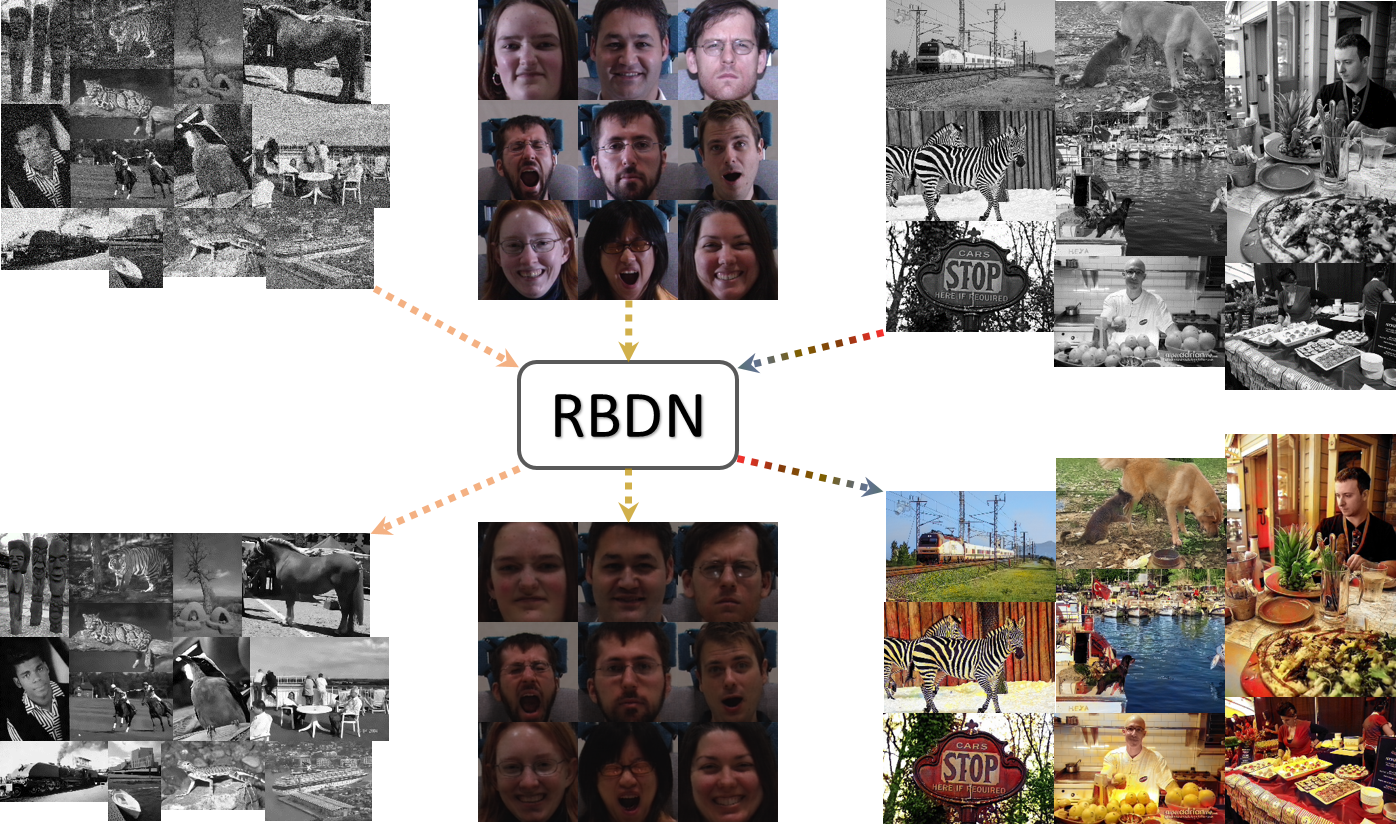}
\end{center}
\caption{Proposed {\bf RBDN} used for diverse Im2Im regression tasks: (from left to right) {\bf Denoising, Relighting, Colorization}.}
\label{fig:placeholder}
\end{figure}

Semantic Segmentation was one of the first applications to witness the extension of DCNNs to output dense pixel wise predictions~\cite{reffcn,refdeconvnet,refdeeplab,reflaplace,refhypercolumns}. These approaches used either VGG or ResNet (without the fully connected layers) as their backbone and introduced architectural changes such as skip layers~\cite{reffcn}, deconvolutional networks~\cite{refdeconvnet,refsegnet}, hypercolumns~\cite{refhypercolumns} or laplacian pyramids~\cite{reflaplace} to facilitate the retention/reconstruction of local input-output correspondences. While these approaches performed very well on segmentation benchmarks, they introduced a trade-off between locality and context. Since the task still remained one of classification (albeit at a pixel level), the trade-off was skewed in favor of incorporating more context and subsequently reconstructing local correspondences from global activations. This is perhaps why some of these approaches had to rely on ancillary methods such as Conditional Random Fields (CRFs)~\cite{refdeconvnet,refdeeplab} to enhance the granularity of their predictions. 

Image-to-Image (Im2Im) regression entails the generation of dense ``continuous" pixel-wise predictions, where the locality-context trade-off is highly task-dependent (typically skewed more in favor of locality). Several DCNN based approaches have been proposed for specific Im2Im regression tasks such as denoising, relighting, colorization, \etc. These approaches typically involve highly task-specific architectures coupled with fine-tuned ancillary post processing methods. However, unlike classification DCNNs, no truly generic architecture for Im2Im regression has yet been proposed which performs consistently well on a diverse range of tasks. It is perhaps the task-dependent locality-context trade-off coupled with the habitual trend of incorporating VGG/ResNet architectures for non-classification tasks, that have impeded progress in this regard.

We propose a generic Im2Im DCNN architecture: {\bf RBDN} which eliminates this trade-off and automatically learns how much locality/context is needed based on the task at hand, through the \emph{early development} of a cheaply computed rich multi-scale image representation using recursive multi-scale branches, learnable upsampling and extensive parameter sharing. 

\begin{figure*}[htpb]
\begin{center}
\includegraphics[width=0.95\linewidth]{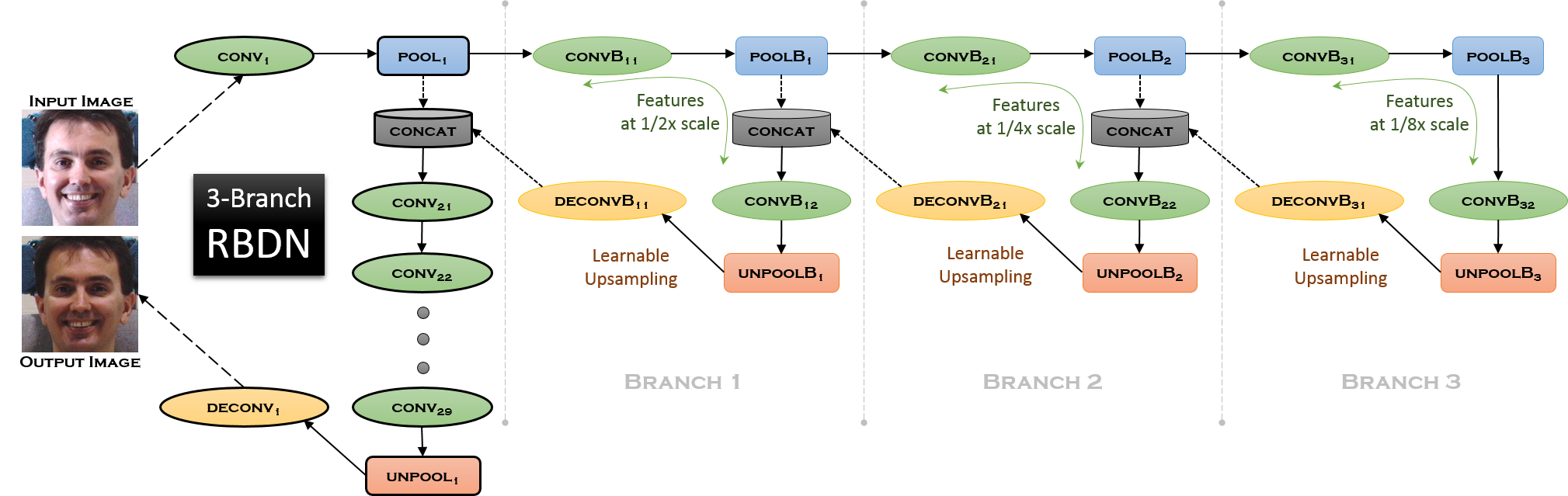}
\end{center}
\caption{Architecture of proposed generic RBDN approach with $3$ branches. The various branches extract features at multiple scales. Learnable upsampling with efficient parameter sharing is used to recursively upsample the activations for each branch until it merges with the POOL$1$ output, leading to a cheap multi-context representation of the input. This multi-context map is subjected to series of $9$ convolutions which can supply ample non-linearity and automatically choose how much context is needed based on the task at hand.}
\label{fig:pipeline}
\end{figure*}

\section{Related Work}
\label{related_work}
We first describe two recently proposed Im2Im DCNN approaches~\cite{refdngen, refv2v} which also have a fairly generic architecture and compare the similarities and differences with our proposed RBDN approach. We then describe some of the related work specific to relighting, denoising and colorization.  

\subsection{Generic Im2Im Regression}
\label{rel_gen}
Deep End-2-End Voxel-2-Voxel prediction~\cite{refv2v} proposed a video-to-video regressor for solving $3$ tasks: semantic segmentation, optical flow and colorization. Their architecture consists of a VGG~\cite{refvgg} style network on which they add branches which upsample and merge activations. Unlike Hypercolumns~\cite{refhypercolumns}, they make the upsampling learnable and perform it in a more efficient way with weight sharing. While~\cite{refv2v} use upsampling to recover local correspondences, DnCNN~\cite{refdngen} on the other hand entirely eliminate downsampling and use a simple $18$ layer fully convolutional network with residual connections for handling $3$ tasks: denoising, super-resolution and jpeg-deblocking. Our proposed RBDN architecture can be viewed as a hybrid of~\cite{refdngen,refv2v}. While we do utilize multi-scale activations like~\cite{refv2v}, we do so very early in the network and generate a cheap composite multi-context representation for the image. Subsequently, we pass the composite map to a linear convolution network like~\cite{refdngen}.

\subsection{Face Relighting}
\label{rel_rl}
In the field of Face Recognition/Verification, while most research focuses on extracting illumination-invariant features, \emph{relighting} is the relatively less explored alternative~\cite{refrl2} of directly making illumination corrections/normalizations to an image. Traditional face relighting approaches used the Retinex~\cite{refrlretinex}/Lambertian Reflectance ~\cite{refrllambert} theory and used spherical~\cite{refrl3,refrllambert}/hemispherical~\cite{refrlhsh} harmonics, subspace-based~\cite{refrlsubspace1, refrlsubspace2} or dictionary-based~\cite{refrldic1,refrldic2,refrldic3,refrldic4,refrldic5,refrldic6} illumination corrections. Deep Lambertian Networks~\cite{refrldeeplambert} encoded lambertian models/illumination corrections directly into their network architecture. This however limited the expressive power of the network, particularly due to the strong lambertian assumptions on isotropicity and absence of specular highlights, which seldom hold true for face images. In section~\ref{exp_rl}, we show that it is possible to train a well-performing relighting model without making any lambertian assumptions using our generic RBDN architecture.     

\subsection{Denoising}
\label{rel_dn}
Denoising approaches typically assume an Additive White Gaussian Noise(AWGN) of known/unknown variance. Traditional denoising approaches include ClusteringSR~\cite{refdnclustsr}, EPLL~\cite{refdnepll}, BM3D~\cite{refdnbm3d}, NL-Bayes~\cite{refdnnlbayes}, NCSR~\cite{refdnncsr}, WNNM~\cite{refdnwnnm}. Among these, BM3D~\cite{refdnbm3d} is the most popular, very well engineered and still widely used as the state-of-the-art denoising approach. Early DCNN based denoising approaches~\cite{refdncnn1,refdncnn2,refdnmlp,refdncnn3,refdncnn4} required a different model to be trained for each noise variance, which limited their practical use. Recently, a Gaussian-CRF based DCNN approach (DCGRF~\cite{refdndcgrf}) was proposed which could explicitly model the noise variance. DCGRF could however only reliably model noise levels within a reasonable range and had to use two models: low-noise DCGRF ($\sigma<25$) and high-noise DCGRF ($25\leq\sigma\leq50$). In section~\ref{exp_dn}, we show that a single model of our proposed RBDN approach trained on a wide range of noise levels ($\sigma\leq50$) achieves competitive results and outperforms all the previously proposed approaches at all noise levels $\sigma\in[25,55]$.

\subsection{Colorization}
\label{rel_color}
The inherent color ambiguity in a majority of objects makes colorization a very hard and ill-posed problem. Early works on colorization~\cite{refcol1, refcol2, refcol3, refcol4, refcol5, refcol6, refcol7, refcol8} required a reference color image from which the color of local patches in the input image was inferred through parametric/non-parametric approaches. Only recently, have DCNN approaches~\cite{refcoldcnn1,refcoldcnn2,refcoldcnn3,refcoldcnn4} been used to solve colorization as an Im2Im classification/regression problem from grayscale to color without requiring auxiliary inputs.~\cite{refcoldcnn1,refcoldcnn3} use Hypercolumns~\cite{refhypercolumns}, while~\cite{refcoldcnn2} use a complex dual-stream architecture that simultaneously identifies/classifies object classes within the image and uses class labels to colorize the input greyscale image. The classification branch of their network is identical to VGG~\cite{refvgg}, while the colorization branch of their network mimics the DeconvNet~\cite{refdeconvnet} architecture. The best colorization results however are obtained by~\cite{refcoldcnn4} despite using a fairly simple VGG~\cite{refvgg} style architecture with dilated convolutions. The key contribution of~\cite{refcoldcnn4} is their novel classification-based loss function over the quantized probability distribution of \emph{ab} values in the Lab color space. They further add a class re-balancing scheme that pushes the predictions away from the statistically likely gray colors, resulting in very colorful colorizations. In section~\ref{exp_color}, we use the same loss function as~\cite{refcoldcnn4} but replace their VGG-style architecture with our proposed RBDN architecture and obtain excellent colorizations.

\section{Generic Im2Im DCNNs}
\label{trans_im2im}
Many Im2Im approaches use VGG/ResNet as their backbone because of their effectiveness and availability. However, this leads to suboptimal architectures (\ref{class_cnn_bad}) for these types of tasks because of the inherent bias towards including more context at the expense of sacrificing locality. We instead propose RBDN (\ref{proposed_approach}) which uses recursive branches to obtain a cheap multi-context locality-preserving image representation very early on in the network. In sections~\ref{base_b0},~\ref{recursive_branches},~\ref{rbdn_analysis}, we describe our network architecture in more detail and analyze its various components.
\subsection{Classification DCNNs are a bad starting point}
\label{class_cnn_bad}
Classification DCNNs typically contain a multitude of interleaved downsampling layers (max-pooling or strided convolutions) which ultimately squash the image to a 1-D vector. With GPU memory being the major bottleneck for training DCNNs, downsampling layers enable the exploration of very deep architectures while providing a natural translational invariance. However, problems arise when attempting to directly port these networks for Im2Im regression tasks. Design changes are needed for retention/recovery of local correspondences, as these get muddled across channels in the middle layers. Recovery with repeated upsampling is inevitably a lossy process, which is particularly harmful for regression tasks demanding continuous pixel-wise predictions. Alternatively, local correspondences can be retained (\eg skip layers, hypercolumns) by merging activation maps from earlier layers at the penultimate layer. The downside to this approach is that activations from very early layers (which contain the bulk of the local correspondences) have a poor capability to model non-linearity, which limits the overall capacity of the network for modeling localized non-linear transformations. For a DCNN to be successful as a generic Im2Im regressor, it would necessarily need to maintain local pixel-wise features, each of which develop strong global representations across the pipeline while independently preserving local information.

\subsection{Proposed Approach: RBDN}
\label{proposed_approach}
Figure~\ref{fig:pipeline} shows the architecture for our proposed Recursively Branched Deconvolutional Network with three branches. At a high level, the network first extracts features at scales $1$(max-locality), $\frac{1}{2}, \frac{1}{4}, \frac{1}{8}$(max-context) and merges all these activations early on to yield a composite map, which is then subjected to a series of convolutions (non-linear transformation) followed by a deconvolution (reconstruction) to yield the output image. The key feature of this network is the multi-scale composite map and how it is efficiently generated using recursive branching and learnable upsampling. During training, the network has a broad locality-context spectrum to work with early on. The series of convolution layers that follow suit can choose the amount of context based on the task at hand and apply ample non-linearity. This translates to a range of modeling capabilities: anywhere from context-aware regression maps to highly localized non-linear transformations (which were difficult to achieve with previously proposed DCNNs).

Our generic $K$-branch RBDN network has two major components: the main branch $B_0$ (which serves as the backbone of our network) and the recursive branches $(B_1,...,B_K)$ (which serve as the head of the network). 

\begin{figure}[!t]
\begin{center}
\includegraphics[width=0.95\linewidth]{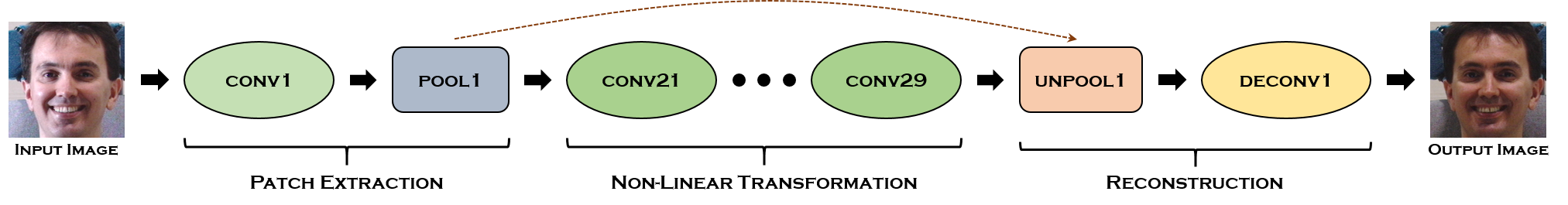}
\end{center}
\caption{Architecture of linear 9-64-3-9 net $B_0$. }
\label{fig:b0_lin}
\end{figure}

\subsection{The Linear Base Network $B_0$}
\label{base_b0}
Inspired by traditional sparse coding approaches, we approach the Im2Im regression problem with a simple network (denoted by its parameters $K$-$c$-$T$-$D$) having three distinct phases:
\begin{itemize}
\item \emph{Patch Extraction}: conv ($K$ x $K$ x $c$) + max-pooling
\item \emph{Non-Linear Transform}: D conv layers ($T$ x $T$ x $c$)
\item \emph{Reconstruction}: unpooling(using max-pool locations) + deconvolution ($K\times K\times c$)
\end{itemize}
We use ReLU~\cite{refrelu} as the activation function and use a batch normalization~\cite{refbn} layer after each convolution/deconvolution. We independently experimented with values $K,c,T,D$ while performing our relighting experiments and found that increasing $K,c,T$ only yields a minor improvement, while increasing the network depth $D$ yielded a significant monotonic improvement until $9$ convolution layers, after which performance saturated. Our final network that gave the best results is shown in figure~\ref{fig:b0_lin}. We denote this network as $B_0$ from here on. (We will use it as the main branch for all RBDN networks).

\begin{figure}[!t]
\begin{center}
\includegraphics[width=0.95\linewidth]{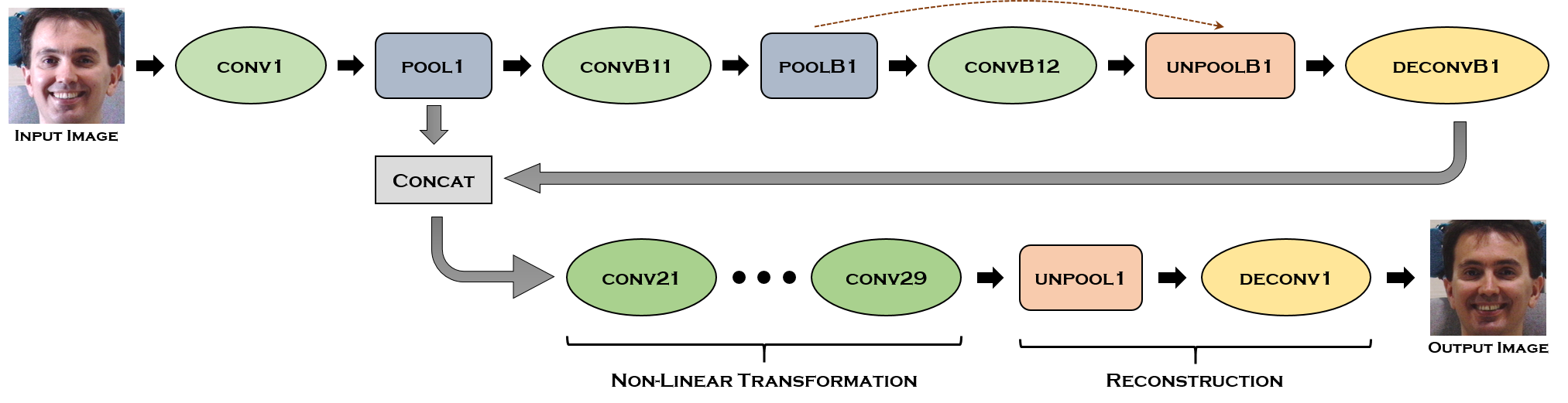}
\end{center}
\caption{Adding the first branch to $B_0$.}
\label{fig:b1_lin}
\end{figure}

\subsection{Recursive Branches $B_0,...B_K$}
\label{recursive_branches}
While the base network $B_0$ by itself gives decent performance for relighting, one of its limitations is a very low field of view. Unlike conventional DCNNs, we cannot add downsampling midway since this would corrupt our local correspondences. As a result, we keep $B_0$ and its local correspondences intact and instead add a branch $B_1$ to the network (see figure~\ref{fig:b1_lin}) at the first pooling layer. Within $B_1$, CONVB$_{11}$+POOLB$_1$+CONVB$_{12}$ computes features at half the scale and UNPOOLB$_1$+DECONVB$_{11}$ provides a learnable upsampling. The output of $B_1$ is then merged with $B_0$ at POOL$1$ itself, which gives the remainder of the network (which invoke the bulk of non-linearity) access to features at $2$ different scales.

\begin{figure}[t!]
\begin{center}
\includegraphics[width=0.95\linewidth]{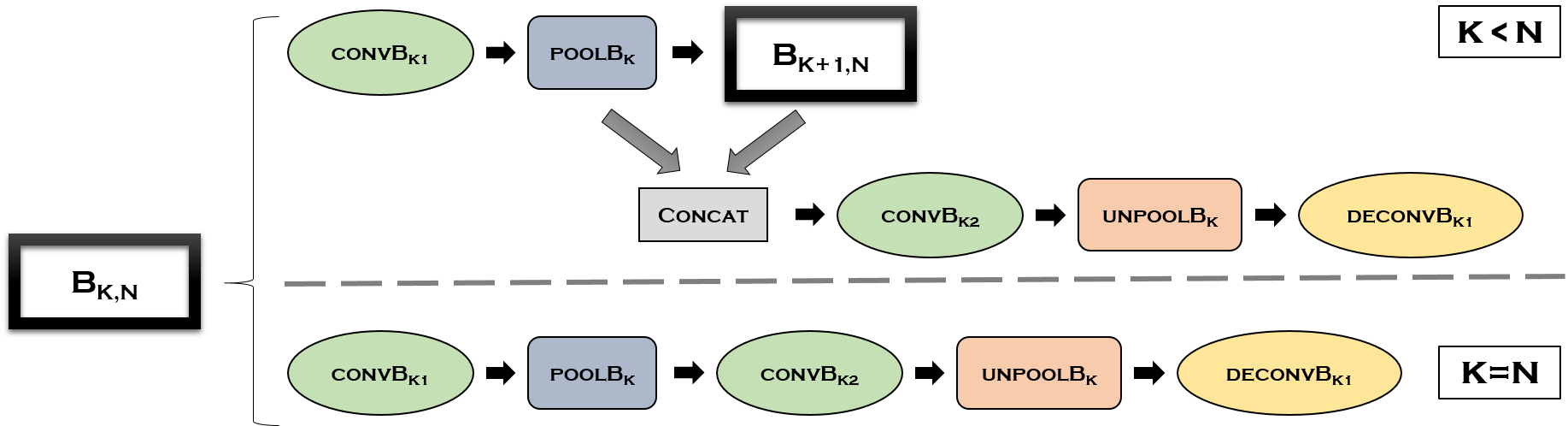}
\end{center}
\caption{Defining the recursive branch module $B_{K,N}$. In the top half, the box with the thick black border, $B_{K+1,N}$ contains the recursive branch. The bottom half of the figure shows the base case (the last branch that does not contain any recursion).}
\label{fig:bk_module}
\end{figure}

We can generalize $B_1$ to multiple branches $B_1,...B_K$. In order to do so, we start by defining the recursive branch module $B_{K,N}$ in figure~\ref{fig:bk_module} which corresponds to the $K^{th}$ branch in a $N$-branch network. Note that branch $B_{K+1,N}$ originates and merges within branch $B_{K,N}$. The advantage of such a recursive construction is two-fold:
\begin{itemize}
\item Activations from deeper branches would have to be upsampled many times before merging with the main branch. The recursive construction helps deeper branches partially benefit from the learnable upsampling machinery in the shallow branches. 
\item Aside from the benefit of parameter sharing, the recursive construction forces activations from deeper branches to traverse a longer path, thus accruing many ReLU activations. This enables deeper branches to model more non-linearity, which is beneficial since they cover a larger Field of View and correspond to global features. 
\end{itemize}

\section{Experiments}
\label{experiments}
We train our generic RBDN architecture for three diverse tasks: relighting, denoising and colorization. We train all our models on a Nvidia Titan-X GPU and use the Caffe~\cite{refcaffe} deep learning framework. For our denoising/colorization experiments, we augment Caffe with utility layers for noise policies (adding WGN to input with $\sigma$ randomly chosen within a user specified range) and image conversions (RGB to YCbCr/Lab space), which streamline the training procedure and enable the use of practically any image dataset out of the box without any pre-processing. We use ReLU~\cite{refrelu} as the activation function and perform Batch Normalization~\cite{refbn} after every convolution/deconvolution layer in all RBDN models.

Unless otherwise mentioned, we train our RBDN models with the mean square error (MSE) as the loss function, crop size of $128$ (chosen randomly from the full-sized training images without any resizing), learning rate of $1$e-$7$, mini-batch size of $64$, step-size of $100000$ and train our model for $500000$ iterations using Stochastic Gradient Descent~\cite{refsgd} (SGD) with momentum and weight decay. During inference, the network by virtue of being fully convolutional can handle variable sized inputs. 

\subsection{Training Datasets}
{\bf CMU-MultiPIE~\cite{refcmumultipie}:} Face images of $337$ subjects are recorded over $4$ sessions. Within a session, there are face images of each subject exhibiting $13$ pose x $19$ illumination x $2$-$3$ expression variations. We used images of $208$ subjects which did not appear in all sessions for training our relighting RBDN, and images of $64$ other subjects for validation. 

{\bf ImageNet ILSVRC2012~\cite{refimagenet}:} $1.2$ million training images and $150,000$ images each for validation and test.

{\bf MS-COCO~\cite{refmscoco}:} $80,000$ training images and $40,000$ images each for validation and test.

For training both our denoising/colorization RBDN, we fuse the train/validation sets of both ImageNet and MS-COCO (total of $1.47$ million training images).

\subsection{Face Relighting}
\label{exp_rl}
We train our relighting RBDN on $20786$ images from CMU-MultiPIE, which takes as input a frontal face image with varying illumination and outputs the image with only ambient lighting. We used a crop size of $224$, step-size of $12000$ and trained our model for $40000$ iterations.

\begin{figure}[!t]
\begin{center}
\includegraphics[width=0.95\linewidth]{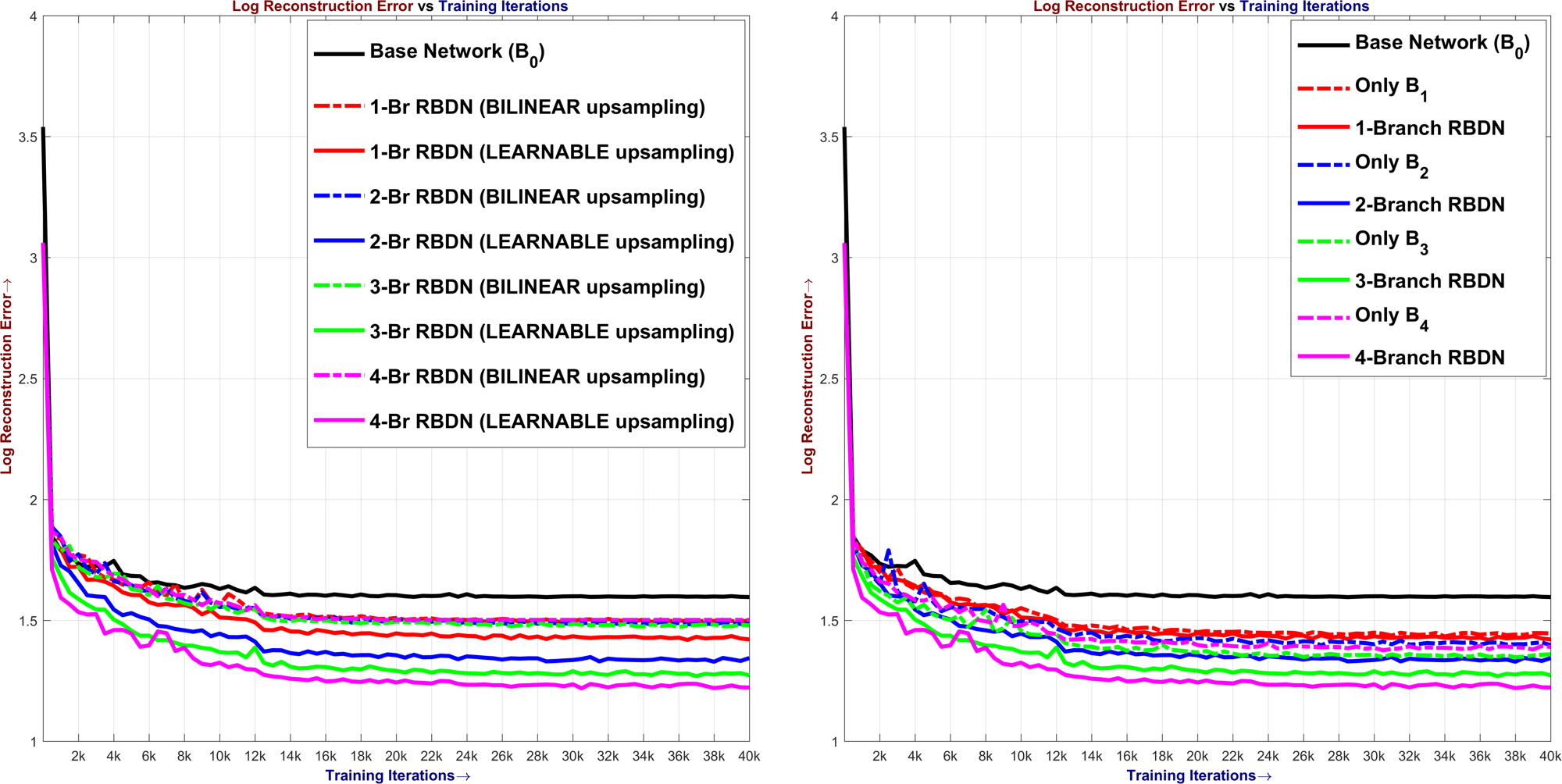}
\end{center}
\caption{Analysing the effect of learnable upsampling(left) and recursive branching(right). Error plots on the CMU-MultiPIE~\cite{refcmumultipie} validation set show a positive influence for both learnable upsampling and recursive branching.}
\label{fig:rbdn_analysis}
\end{figure} 

\subsubsection{Analysis of RBDN} 
\label{rbdn_analysis}
Compared to the base network $B_0$, a $K$-branch RBDN has two major additions: the recursive branching and learnable upsampling. We perform two sets of relighting experiments to independently observe the efficacy of both on a $K$-branch RBDN($K=0,1,2,3,4$) as follows:  
\begin{itemize}
\item We removed the CONCAT layers which merge the different branches. This resulted in a linear network ($B_K$ only) similar in structure to the deconvolutional networks used for semantic segmentation~\cite{refdeconvnet,refsegnet}.
\item We replaced the learnable upsampling with fixed bilinear upsampling. 
\end{itemize}
Figure~\ref{fig:rbdn_analysis} shows the error plots of log reconstruction error on the CMU-MultiPIE~\cite{refcmumultipie} validation set vs training iterations for both experiments. The plots show that both learnable upsampling and recursive branching independently have a positive impact on performance.

\begin{figure}[t!]
\begin{center}
\includegraphics[width=\linewidth]{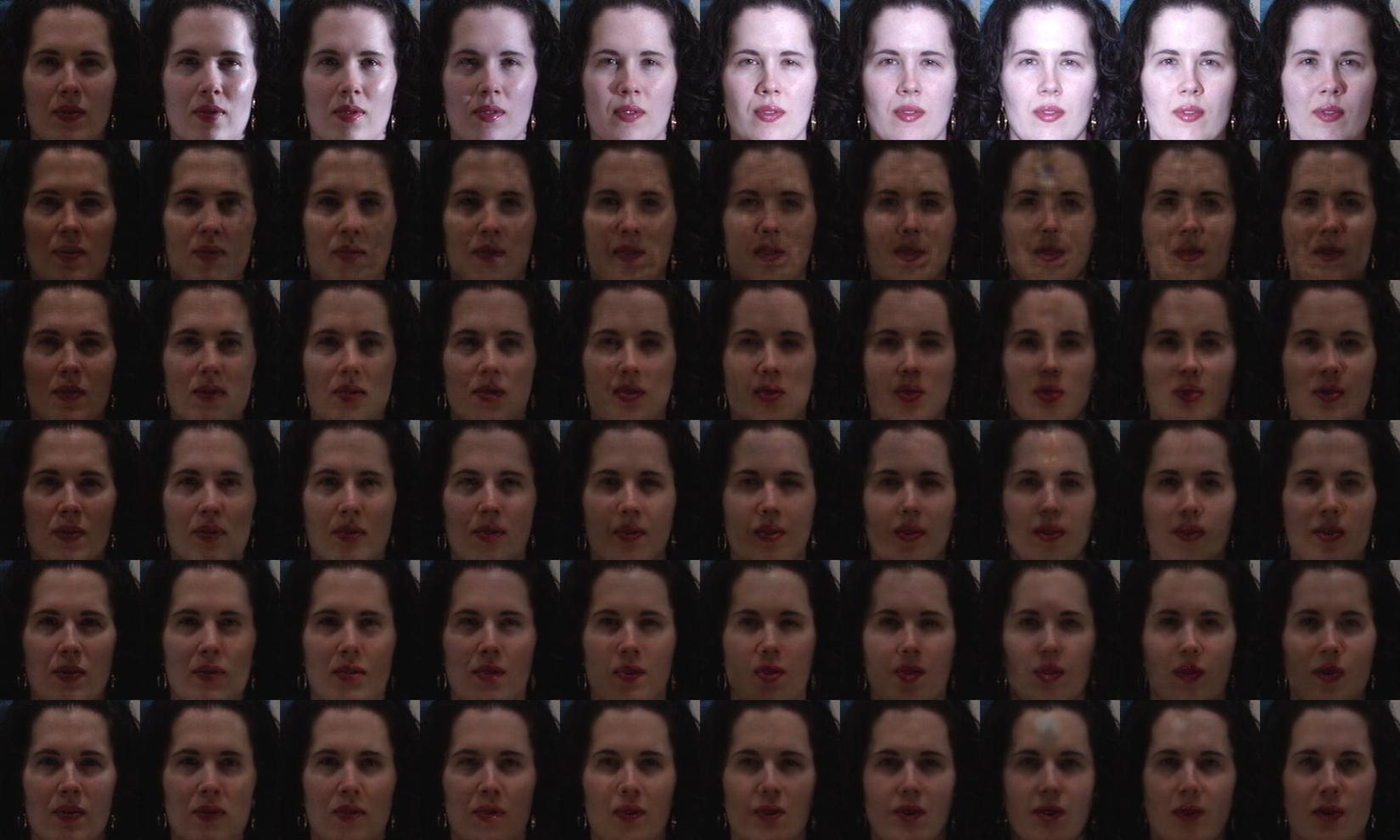}
\end{center}
\caption{Relighting RBDN results for a subject from the CMU-MultiPIE~\cite{refcmumultipie} validation set. {\bf Top Row:} Input images (ground truth is top-left image). {\bf $2^{nd}$ row:} $B_0$ output (no branches; strong artifacts can be seen.) {\bf $3^{rd}$-$6^{th}$ row:} RBDN outputs for $1,2,3,4$ branches respectively. Results improve with increase in number of branches up to $3$ branches. The network starts overfitting at $4$ branches.}
\label{fig:multipie1}
\end{figure} 

\begin{figure*}[htpb]
\begin{center}
\includegraphics[width=0.95\linewidth]{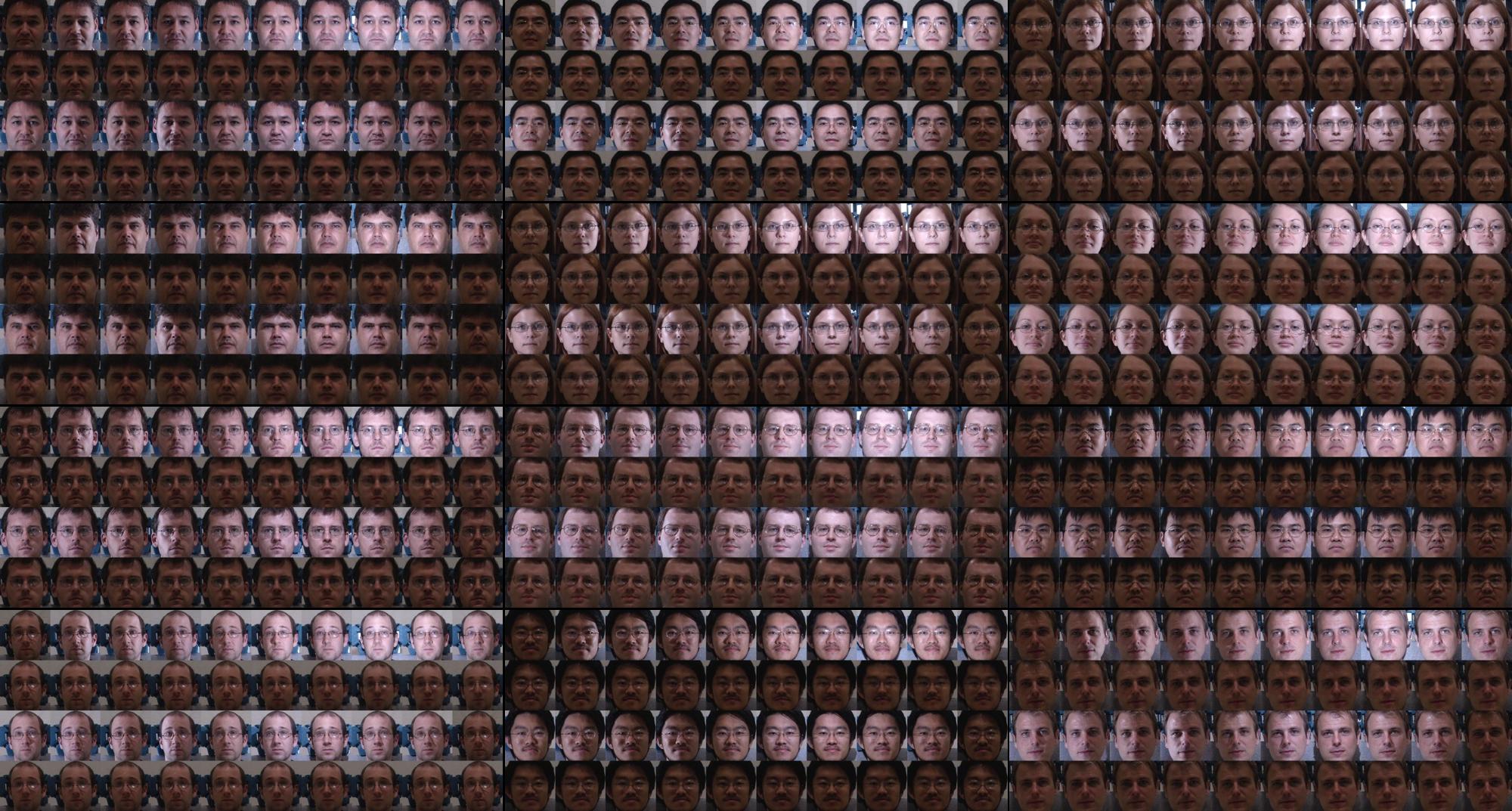}
\end{center}
\caption{Relighting results on the CMU-MultiPIE validation set. The goal is to render faces from various unknown lighting conditions to a fixed lighting condition. {\bf Odd rows:} Inputs (top-left image for each subject is the ground truth), {\bf Even Rows}: $3$-branch RBDN output}
\label{fig:multipie2a}
\end{figure*}

\begin{table*}[htpb]
\centering
\resizebox{\textwidth}{!}{%
\begin{tabular}{@{}cccccccccccc@{}}
\toprule
Test $\sigma$                & 10                                 & 15                                 & 20                                 & 25                                 & 30                                 & 35                                 & 40                                 & 45                                 & 50                                 & 55                                 & 60                                 \\ \midrule
ClusteringSR~\cite{refdnclustsr}                 & 33.27                              & 30.97                              & 29.41                              & 28.22                              & 27.25                              & 26.30                              & 25.56                              & 24.89                              & 24.28                              & 23.72                              & 23.21                              \\
EPLL~\cite{refdnepll}                         & 33.32                              & 31.06                              & 29.52                              & 28.34                              & 27.36                              & 26.52                              & 25.76                              & 25.08                              & 24.44                              & 23.84                              & 23.27                              \\
BM3D~\cite{refdnbm3d}                         & 33.38                              & 31.09                              & 29.53                              & 28.36                              & 27.42                              & 26.64                              & 25.92                              & 25.19                              & 24.63                              & 24.11                              & 23.62                              \\
NL-Bayes~\cite{refdnnlbayes}                     & 33.46                              & 31.11                              & 29.63                              & 28.41                              & 27.42                              & 26.57                              & 25.76                              & 25.05                              & 24.39                              & 23.77                              & 23.18                              \\
NCSR~\cite{refdnncsr}                         & 33.45                              & 31.20                              & 29.56                              & 28.39                              & 27.45                              & 26.32                              & 25.59                              & 24.94                              & 24.35                              & 23.85                              & 23.38                              \\
WNNM~\cite{refdnwnnm}                         & \textbf{33.57}                              & 31.28                              & 29.70                              & 28.50                              & 27.51                              & 26.67                              & 25.92                              & 25.22                              & 24.60                              & 24.01                              & 23.45                              \\ \midrule
TRD~\cite{refdntrd}                          & -                                  & 31.28                              & -                                  & 28.56                              & -                                  & -                                  & -                                  & -                                  & -                                  & -                                  & -                                  \\
MLP~\cite{refdnmlp}                          & 33.43                              & -                                  & -                                  & 28.68                              & -                                  & 27.13                              & -                                  & -                                  & 25.33                              & -                                  & -                                  \\ \midrule
DCGRF~\cite{refdndcgrf}            & 33.56 & \textbf{31.35} & \textbf{29.84} & 28.67          & 27.80          & 27.08          & 26.44          & 25.88          & 25.38          & 24.90          & \textbf{24.45}                              \\ \midrule
DnCNN~\cite{refdngen}   & 33.32     & 31.29 & \textbf{29.84} & 28.68 & 27.70  &  26.84  & 26.05  & 25.34  & 24.68  & 24.05  & 23.39 \\ \midrule
\textbf{3-branch RBDN}          & 32.85          & 31.05          & 29.76          & \textbf{28.77} & \textbf{27.97} & \textbf{27.31} & \textbf{26.73} & \textbf{26.24} & \textbf{25.80} & \textbf{25.22} & 23.25 \\ \bottomrule

\end{tabular}
}
\caption{Mean PSNR for various denoising approaches on $300$ test images. A \emph{single} denoising model is used to report all results for {\bf RBDN} (trained on $\sigma\in[8,50]$) and DnCNN~\cite{refdngen} (trained on $\sigma\in[0,55]$). For other comparison approaches, note that the best performing model at each noise level is used to report results.}
\label{tabdenoising}
\end{table*}

\begin{figure*}[htpb]
\begin{center}
\includegraphics[width=\linewidth]{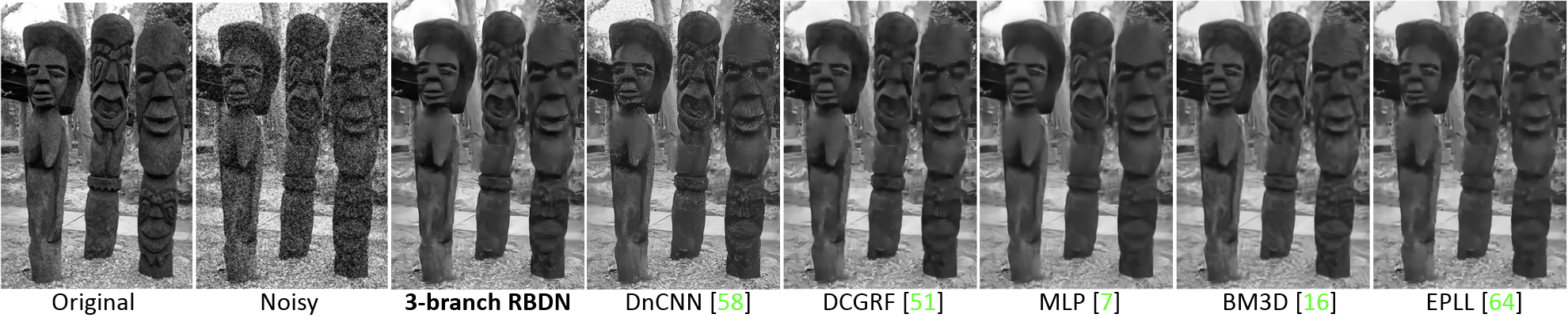}
\end{center}
\caption{Visual comparison of various denoising approaches on a test image from BSD300 with WGN of $\sigma=50$.}
\label{fig:dn1comp}
\end{figure*}

\begin{figure*}[htpb]
\begin{center}
\includegraphics[width=\linewidth]{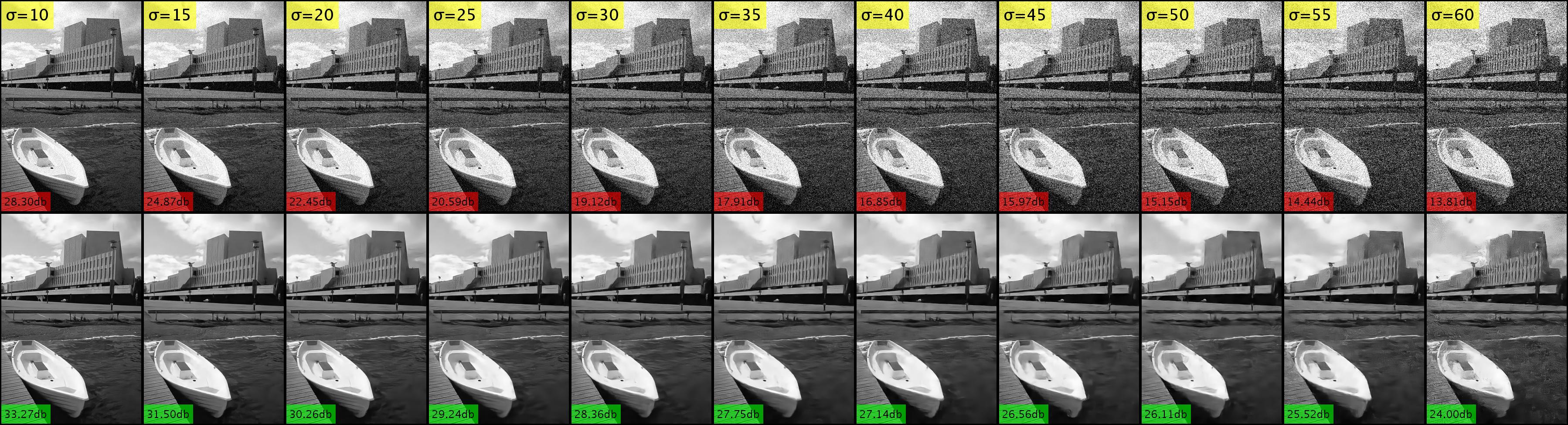}
\end{center}
\caption{Illustrating the capability of a single RBDN model to handle a range of noise levels(yellow box). {\bf Top Row:} Noisy test image (PSNR in red box). {\bf Bottom Row:} Denoised result with $3-$branch RBDN (PSNR in green box)}
\label{fig:dn_allnoise}
\end{figure*}

\begin{figure*}[htpb]
\begin{center}
\includegraphics[width=\linewidth]{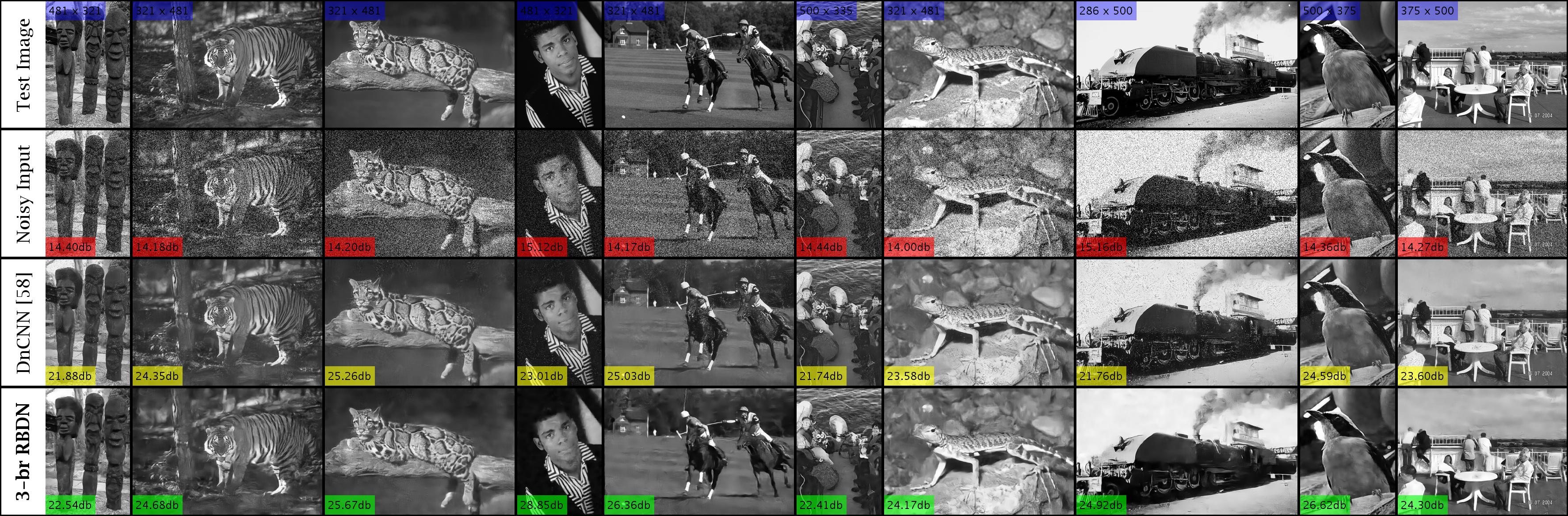}
\end{center}
\caption{Illustrating RBDN's ability to reliably denoise at $\sigma=55$, outside our training bounds ($\sigma\in[8,50]$). The $18$-layer DnCNN~\cite{refdngen} (despite using $\sigma=55$ for training) is outperformed by our $9$-layer RBDN. Red, Yellow, Green boxes show the PSNR.}
\label{fig:dn55a}
\end{figure*}

\subsection{Denoising}
\label{exp_dn}
We train a single $3$-branch RBDN model for denoising which takes as input a grayscale image corrupted by additive WGN with standard deviation uniformly randomly chosen in the range $\sigma\in[8,50]$. We use the same evaluation protocol as ~\cite{refdndcgrf}, with a $300$ image test set (all $100$ images of the BSD300~\cite{refbsd300} test set and $200$ images from PASCAL VOC2012~\cite{refpascalvoc} dataset). Precomputed noisy test images from~\cite{refdndcgrf}, that are quantized to the $[0,255]$ range are used to compare various approaches for a fair realistic evaluation.

\begin{figure*}[htpb]
\begin{center}
\includegraphics[width=\linewidth]{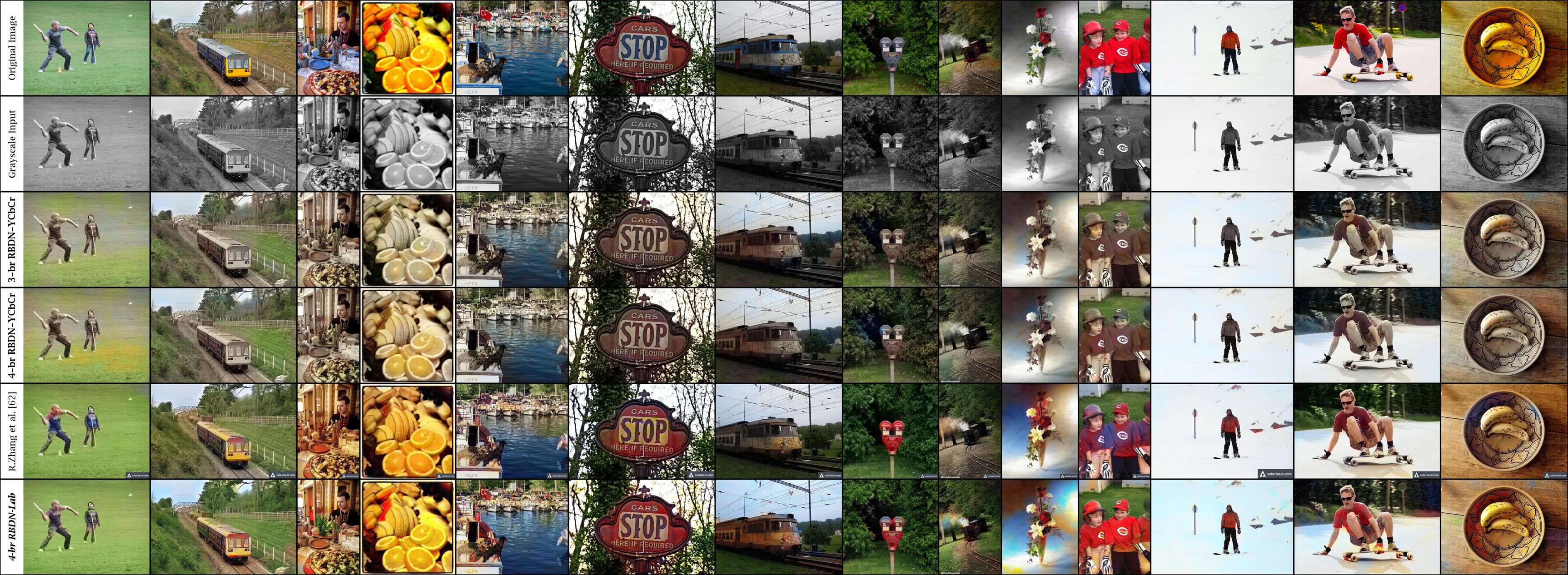}
\end{center}
\caption{Colorization results for images from MS-COCO test set.}
\label{fig:comp1}
\end{figure*}

\subsection{Colorization}
\label{exp_color}
We first transform a color image into YCbCr color space and predict the chroma (Cb,Cr) channels from the luminance (Y-channel) input using RBDN. The input Y-channel is then combined with the predicted Cb,Cr channels and converted back to RGB to yield the predicted color image. We denote this model as {\bf RBDN-YCbCr}. 

Inspired by the recently proposed Colorful Colorizations~\cite{refcolorzhang} approach, we train another RBDN model which takes as input the L-channel of a color image in \emph{Lab} space and tries to predict a $313$-dimensional vector of probabilities for each pixel (corresponding to $313$ \emph{ab} pairs resulting from quantizing the \emph{ab}-space with a grid-size of $10$). Subsequently, the problem is treated as multinomial classification and we use a softmax-cross-entropy loss with class re-balancing as in~\cite{refcolorzhang}. Instead of SGD, we use the Adam~\cite{refsolveradam} solver for training, with a learning rate of $3.16$e-$3$ ($\gamma=0.316$), step-size of $45000$, mini-batch size of $128$ and train our model for $200000$ iterations. During inference, we use the annealed-mean of the softmax distribution to obtain the predicted \emph{ab}-channels as in~\cite{refcolorzhang}. We denote this model as {\bf RBDN-Lab}.

\section{Results}
\label{results}
{\bf Relighting:} Figure~\ref{fig:multipie1} shows the RBDN outputs with $0,1,2,3,4$ branches for a subject from the CMU-MultiPIE validation set. The improvement in results from $B_0$ (no branches) to $1$-branch RBDN is very prominent, after which there is a gradual improvement with increase in number of branches up to $3$. Results deteriorate when transitioning to a $4$-branch RBDN (possibly due to overfitting on the relatively small dataset). Figure~\ref{fig:multipie2a} shows some more results from the validation set for the $3$-branch RBDN, which achieves near perfect relighting for all subjects.

{\bf Denoising:} Table~\ref{tabdenoising} shows the mean PSNR for various denoising approaches on the $300$ benchmark test images. Besides RBDN, DnCNN~\cite{refdngen} and DCGRF~\cite{refdndcgrf}, all other approaches train a separate model for each noise level. For DCGRF~\cite{refdndcgrf}, results are reported with a low noise model for test $\sigma\leq25$ and a high noise model for test $\sigma\geq30$. The results for both DnCNN~\cite{refdngen} and our $3$-branch RBDN however correspond to a \emph{single} model trained to automatically handle \emph{all} noise levels. Our model outperforms all the other approaches at test noise $\sigma\in[25,55]$. Figure~\ref{fig:dn1comp} shows a visual comparison of various denoising approaches for a test image from BSD300. Figure~\ref{fig:dn_allnoise} highlights a single RBDN model's denoising capability across a range of noise levels. Figure~\ref{fig:dn55a} illustrates the generalization ability of the RBDN to reliably denoise at a very high noise level of $\sigma=55$ (which is outside the bounds of our training). The fact that our $9$-layer RBDN (without any residual connections~\cite{refresnet}) outperforms the $18$-layer residual DnCNN~\cite{refdngen}, suggests that cheap early recursive branching is more beneficial than added depth. 

{\bf Colorization:} Figure~\ref{fig:comp1} shows the colorizations of various models on the MS-COCO test set. The $3,4$-branch RBDN-YCbCr models produce decent colorizations, but are very dull and highly under-saturated. This is however not an architectural limitation, but rather the MSE loss function which tends to push results towards the average. Colorization is inherently ambiguous for a large majority of objects such as cars, people, animals, doors, utensils, \etc, several of which can take on a wide range of permissible colors.  On the other hand, the MSE based models are able to reasonably color grass, sky, water as these typically take on a fixed range of colors. Softmax cross-entropy loss based models with class rebalancing (\cite{refcolorzhang} and the $4$-branch RBDN-Lab) are able to overcome the under-saturation problem by posing the problem as a classification task and forcibly pushing results away from the average. Finally, the only difference between the $4$-branch RBDN-Lab and the linear dilated convolutional network of ~\cite{refcolorzhang} is the architecture. Both models give very good colorizations, with one appearing better than the other for certain images and vice-versa, although the colorizations of RBDN-Lab have a higher saturation and appear slightly more colorful for all images.     

\section{Conclusion and Future Work}
\label{conclusion}
We proposed a DCNN architecture for Im2Im regression: RBDN, which gives competitive results on $3$ diverse tasks: relighting, denoising and colorization, when used off-the-shelf without any task-specific architectural modifications. The key feature of RBDN is the development of a cheap multi-context image representation early on in the network, by means of recursive branching and learnable upsampling, which alleviates the locality-context trade-off concerns inherent in the design of Im2Im DCNNs. 

We believe that several improvements can be made to the RBDN architecture. First, the RBDN architecture could potentially benefit from residual connections, dilated convolutions and possibly other activation functions besides ReLU. Secondly, we used a network of fixed depth across all tasks, which may prove insufficient for complex tasks or suboptimal for simple tasks. The recently proposed Structured Sparsity approach~\cite{refstructsparse} allows networks to simultaneously optimize their hyperparameters (filter size, depth, local connectivity) in a highly efficient way while training by means of Group Lasso~\cite{refgrouplasso} regularization. Thirdly, MSE is known to be an extremely poor~\cite{refmsepoor} loss function for tasks demanding perceptually pleasing image outputs. While the loss function from ~\cite{refcolorzhang} we used for colorization overcame MSE's limitations, it is specific to the colorization problem. Loss functions based on Adversarial Networks~\cite{refgan} on the other hand can be a generic MSE replacement.

\section{Acknowledgements}
This research is based upon work supported by the Office
of the Director of National Intelligence (ODNI), Intelligence
Advanced Research Projects Activity (IARPA), via
IARPA R\&D Contract No. 2014-14071600012. The views
and conclusions contained herein are those of the authors
and should not be interpreted as necessarily representing the
official policies or endorsements, either expressed or implied,
of the ODNI, IARPA, or the U.S. Government. The U.S.
Government is authorized to reproduce and distribute reprints
for Governmental purposes notwithstanding any copyright
annotation thereon.


{\small
\bibliographystyle{ieee}
\bibliography{venkai_cvpr2017_refs}
}

\end{document}